\documentclass[twocolumn]{article}
\usepackage{times}
\usepackage{soul}
\usepackage{url}
\usepackage[hidelinks]{hyperref}
\usepackage[utf8]{inputenc}
\usepackage[small]{caption}
\usepackage{graphicx}
\usepackage{amsmath}
\usepackage{amsthm}
\usepackage{amsfonts}
\usepackage{booktabs}
\usepackage{algorithm}
\usepackage{algorithmic}
\usepackage{authblk}
\usepackage{bm}
\newcommand{\bA}{{\bm A}}
\newcommand{\bH}{{\bm H}}

\newcommand{\bW}{{\bm W}}
\newcommand{\bI}{{\bm I}}
\newcommand{\bF}{{\bm F}}
\newcommand{\bE}{{\bm E}}
\newcommand{\bx}{{\bm x}}
\newcommand{\be}{{\bm e}}

\title{End-to-End Entity Classification on Multimodal Knowledge Graphs}

\author[1]{W.X.~Wilcke\thanks{w.x.wilcke@vu.nl}}
\author[1]{P.~Bloem\thanks{p.bloem@vu.nl}}
\author[1]{V.~de~Boer\thanks{v.de.boer@vu.nl}}
\author[2]{R.H.~van~'t~Veer\thanks{rein.van.t.veer@geodan.nl}}
\author[1]{F.A.H.~van~Harmelen\thanks{frank.van.harmelen@vu.nl}}
\affil[1]{Department of Computer Science, Vrije Universiteit, Amsterdam, The Netherlands}
\affil[2]{Geodan, Amsterdam, The Netherlands}

\begin{document}

\maketitle

\begin{abstract}	
End-to-end multimodal learning on knowledge graphs has been left largely unaddressed. Instead, most end-to-end
models such as message passing networks learn solely from the relational information encoded in graphs' structure: raw
values, or literals, are either omitted completely or are stripped from their values and treated as regular nodes.
In either case we lose potentially relevant information which could have otherwise been exploited by our learning
methods. To avoid this, we must treat literals and non-literals as separate cases. We must also address each
modality separately and accordingly: numbers, texts, images, geometries, et cetera. We propose a multimodal message
passing network which not only learns end-to-end from the structure of graphs, but also from their possibly divers set
of multimodal node features. Our model uses dedicated (neural) encoders to naturally learn embeddings for node
features belonging to five different types of modalities, including images and geometries, which are projected into a joint
representation space together with their relational information. We demonstrate our model on a node classification
task, and evaluate the effect that each modality has on the overall performance. Our result supports our hypothesis
that including information from multiple modalities can help our models obtain a better overall performance.
\end{abstract}

\section{Introduction}

\begin{figure}[t]
  \includegraphics[width=\linewidth]{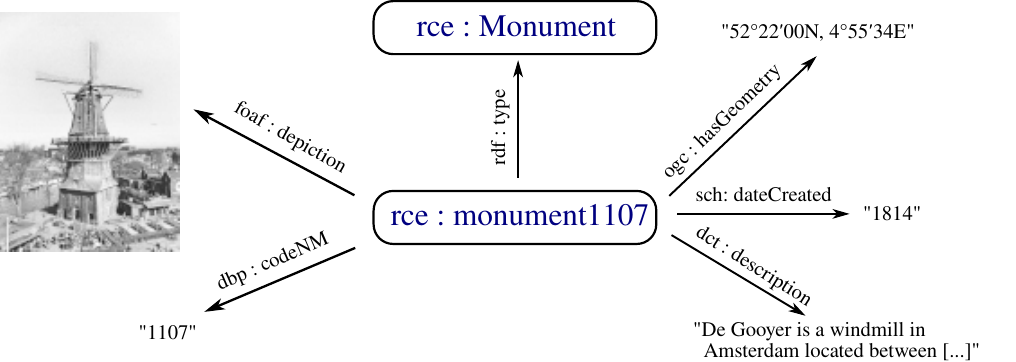}
  \caption{A simplified and incomplete example from the Dutch Monuments Graph showing a single monument with several
  attributes of different modalities.}
  \label{fig:examplegraph}
\end{figure}

The recent adoption of knowledge graphs by multinationals such as Google and Facebook has made them interesting targets
for various machine learning applications such as link prediction and node classification. Already, this interest has
lead to the development of message passing models which enable data scientists to learn end-to-end\footnotemark from any
arbitrary graph. To do so, these models exploit the relational information encoded in the graphs' structure to guide the
learning process. The same approach has also been shown to work quite well on knowledge graphs, obtaining results that
are comparable to dedicated models such as \textit{RDF2Vec}~\cite{ristoski2016collection} and Weisfeiler-Lehman
kernels~\cite{shervashidze2011weisfeiler}. Nevertheless, by focusing on a single modality---the graphs' structure---we
are effectively throwing away a lot of other information that knowledge graphs tend to have, and which, if we were able
to include them in the learning process, have the potential of improving the overall performance of our models.

\footnotetext{In the context of this paper, we define ``end-to-end learning'' as the use of machine learning models
which operate directly on raw data, instead of relying on manually engineered features. In end-to-end learning, any
information in the data can, in principle, be used by the model. See \cite{wilcke2017the} for a more in-depth
discussion.}

Combining information from multiple modalities is a topic that is already well studied for information stored in
\emph{relational} form (for instance in relational database management systems). Here too, we often encounter
\emph{heterogeneous} knowledge, containing information from a wide variety of modalities (such as language, audio, or
images). In \cite{wilcke2017the}, the case is made that to truly learn \emph{end-to-end} from a collection of
heterogeneous, multimodal data, we must design machine learning models that can consume these data in as raw a form as
possible, staying as close as we can to the original knowledge, and that we need to adopt a data model which can
represent our data in a suitable format, for which the knowledge graph is a natural choice. In other words, even when
our heterogeneous multimodal data is not initially represented as a knowledge graph, transforming it to this format is a
natural first step in an end-to-end multimodal machine learning pipeline. 

In this paper, we aim to show a first proof-of-concept model for this principle by introducing a message passing neural
network which can directly consume heterogeneous multimodal data, represented as knowledge graph, and which itself can
learn to extract relevant information from each modality, based solely on the downstream task.

We call a knowledge graph that contains information in multiple modalities a \emph{multimodal knowledge graph}. The
most elementary modality---the relational information---is encoded in the graph structure. Other common modalities are
of numerical, textual, and temporal nature, such as various measurements, names, and dates, respectively, and, in a
lesser degree, of visual, auditory, and spatial makeup. In a knowledge graph about monuments, for example, we might find
that each monument has a detailed description, a registration number, a year in which it was build, a few pictures from
different angles, and a set of coordinates (Figure~\ref{fig:examplegraph}). These and other attributes are encoded as
raw values with a corresponding datatype declaration, called \textit{literals}, and tell us something about the objects
they are connected to, called \textit{entities}. However, most of this information is lost when we reduce the literals
to identifiers, as is currently common practice when we apply message passing networks to knowledge graphs. 

By reducing literals to identifiers, we discard any information that is contained in their contents, retaining only the
relational information encoded by their connections, and placing them on an equal footing with all other entities. This
means that we are effectively feeding our models a subset of the original and complete knowledge, but also that we are
depriving our models of the ability to compare inputs according to their modalities: measurements as numbers,
descriptions as language, coordinates as geometries, etc. As a result, our models are unable to distinguish between literals
that are closely together in the value space with those which are far apart. The name \textit{Mary}, for example, would
be seen as (dis)similar to \textit{Maria} as it would to \textit{Wilberforce}, as would the integer value $\mathit{47}$
be to $\mathit{42}$ and $\mathit{6.626068 \times 10^{-34}}$. Instead however, we want our models to use this information
to guide their learning process. 

By enabling our models to naturally ingest literal values, and by treating these values according to their modalities,
tailoring their encodings to their specific characteristics, we stay much closer to the original and complete
knowledge that is available to us. We believe that doing so enables our models to create better internal
representations of the entities we are trying to learn over, potentially resulting in an increase in the overall
performance of our models. In this work, we test this supposition by feeding information from many different modalities
through dedicated (neural) encoders into a joint representation space by means of late fusion. By embedding our
approach within the message passing framework, and by exploiting datatype declarations and common vocabularies such as
\textit{XSD}\footnote{XML Schema Definition Language} and \textit{OGC}\footnote{Open Geospatial Consortium}, we embrace
the idea that this enables us to learn end-to-end from any heterogeneous multimodal data, as long as they
are represented as knowledge graph. To evaluate our supposition, we investigate the influence of each separate modality on the
classification accuracy on six different heterogeneous multimodal knowledge graphs. 

Because the interest in multimodal learning on knowledge graphs has emerged only recently, only few multimodal benchmark
datasets exist, most of which only include numerical and textual information~\cite{liu2019mmkg,emnlp18}. Images are also
often included, but are stored outside the graph and linked to using hyperlinks and imported during runtime. To add to
this modest collection, we have created a knowledge graph about monumental buildings in the Netherlands that includes
heterogeneous information from six different modalities, all of which are incorporated in the graph (see
Sc.~\ref{sec:datasets} for more details). We offer this dataset to the community with the hope that it can be used to
further the research on this topic.

To summarize, the main contributions of this paper are:
\begin{enumerate}
\item A machine learning model, embedded in the message passing
framework, which can learn end-to-end from any heterogeneous knowledge graph, and which can naturally ingest
literal values according to their modalities. 
\item An inverse ablation study on the potential usefulness of
including information from multiple modalities, and the effect this has on the overall performance of our models. 
\item A knowledge graph about monuments in The Netherlands which contains information from six different modalities, and which we offer as benchmark.
\end{enumerate}

Our aim is emphatically \emph{not} to show that our approach achieves any kind of state-of-the-art, or even to measure its performance against related models. For this purpose the available benchmark data is insufficiently mature.  Rather, we present our approach as a proof-of-concept. We show that in certain cases, a model can be trained end-to-end on a heterogeneous knowledge graph so that it learns purely from the downstream classification task, which patterns to extract from each modality.

\section{Related Work}

Machine learning from multimodal sources is a well-studied problem. A good introduction to the problem and its many
perspectives is given by \cite{baltruvsaitis2018multimodal}. According to their taxonomy, our approach is one of
\emph{late fusion} by message passing network, focusing on \emph{representation} of multiple modalities in a joint
representation space. We consciously ignore the hard problems of \emph{alignment} and \emph{translation}: data in a given
modality is only every used to learn a vector representation of a literal node. 

Various other approaches have explored using the information from literal nodes from one or more modalities in knowledge
graph machine learning models. An overview is provided by \cite{gesese2019survey} for the specific use case of link
prediction. While two of the models surveyed, only MKBE \cite{emnlp18} use literals representing a variety of
modalities, including images. Like our approach, MKBE uses a set of modality specific (neural) encoders to map multimodal
information to embedding vectors.

All these models are simple embedding models, based on a score function applied to triples. By contrast, our approach
includes a message passing layer, allowing multimodal information to be propagated through the graph, several
hops, before being used for classification.

Our model is currently only evaluated on entity classification, putting a direct comparison to these methods out of
scope.

\section{Preliminaries}

Knowledge graphs and message passing neural networks are integral components of our research. We will here briefly introduce both
concepts.

\subsection{Knowledge Graphs} For the purposes of this paper we define a \textit{multimodal knowledge graph} $G = (\mathcal{V}, \mathcal{E})$ over
modalities $1, \ldots, \mathcal{M}$ as a labeled multidigraph defined by a set of nodes $\mathcal{V} = \mathcal{I} \cup
\mathcal{L}^m$ and a set of directed edges $\mathcal{E}$, and with $n = |\mathcal{V}|$. Nodes belong to one of two categories: entities $\mathcal{I}$, which
represent objects (monuments, people, concepts, etc.), and literals $\mathcal{L}^m$, which represent raw values in
modality $m \in \mathcal{M}$ (numbers, strings, coordinates, etc.). We also define a set of relations $\mathcal{R}$
which contains the edge types that make up $\mathcal{E}$. Relations are also called \textit{predicates}.

Information in $G$ is encoded as triples $\mathcal{T}$ of the form $(h, r, t)$, with head $h \in \mathcal{I}$, relation
$r \in \mathcal{R}$, and tail $t \in \mathcal{I} \cup \mathcal{L}^1 \cup \ldots \cup \mathcal{L}^m$. The combination of
relations and literals are also called \textit{attributes} or \textit{node features}.

See Figure~\ref{fig:examplegraph} for an example of knowledge graph with seven nodes, two of which are entities and the
rest literals. All knowledge graphs in this paper are
stored in the \textit{Resource Description Framework} format~\cite{lassila1998resource}, but our model can be applied to
any graph fitting the above definition.

\subsection{Message Passing Neural Networks}
A \emph{message passing neural network} \cite{gilmer2017neural} is a graph neural network
model that uses trainable functions to propagate node embeddings over the edges of the neural network. One simple
approach to message passing is the graph convolutional neural network (GCN)  \cite{kipf2016semi}. The
R-GCN~\cite{schlichtkrull2018modeling}, on which we build, is a straightforward extension to the knowledge graph setting.

Let $\bH^0$ be a $n \times q$ matrix of $q$ dimensional node embeddings for all $n$ nodes in the graph. That is, the
$i$-th row of $\bH^0$ is an embedding for the $i$-th node in the graph~\footnotemark, The R-GCN computes an updated $n
\times l$ matrix $\bH^1$ of $l$-dimensional node embeddings by the following computation (the \emph{graph convolution}):

\begin{equation}
	\label{eq:rgcn}
 \bH^1 = \sigma\left(\sum_{r\in \mathcal{R}} \bA^r\bH^0\bW^r \right)
 \end{equation}

 Here, $\sigma$ is an activation function like
ReLU, applied element-wise. $\bA^r$ is the row-normalised adjacency matrix for the relation $r$ and $\bW^r$ is a $q
\times l$ matrix of learnable weights. This operation arrives at a new node embedding for a node by averaging the
embeddings of all its neighbours, and linearly projecting to $l$ dimensions by $\bW^r$. The embeddings are then summed
over all relations and a non-linearity $\sigma$ is applied. 

\footnotetext{The standard R-GCN does not distinguish between literals and entities. Also, literals with the same value
are collapsed into one node, therefore $n \leq |\mathcal{V}|$.}

To use R-GCNs for entity classification with $c$ classes, the standard approach is to start with one-hot vectors as
initial node embeddings (that is, $\bH^0 = \bI$). These are transformed to $h$-dimensional node embeddings by a first
R-GCN layer (commonly with $h=16$), which are transformed to $c$-dimensional node embeddings by a second R-GCN layer. The
second layer has a row-wise softmax non-linearity, so that the final node embeddings can be read as class probabilities.
The network is then trained by computing the cross-entropy loss for the known labels and backpropagating to update the
weights. Using more than two layers of message passing does not commonly improve performance with current message
passing models.

To allow information to propagate in both directions along an edge, all inverse relations are added to the predicate
set. The identity relation is also added (for which $\bA^r = \bI$) so that the information in the current embedding can,
in principle, be retained. To reduce overfitting, the weights $\bW^r$ can be derived from a smaller set of \emph{basis
weights} by linear combinations (see the original paper for details).

\section{A Multimodal Message Passing Network}

\begin{figure}[t]
  \includegraphics[width=\linewidth]{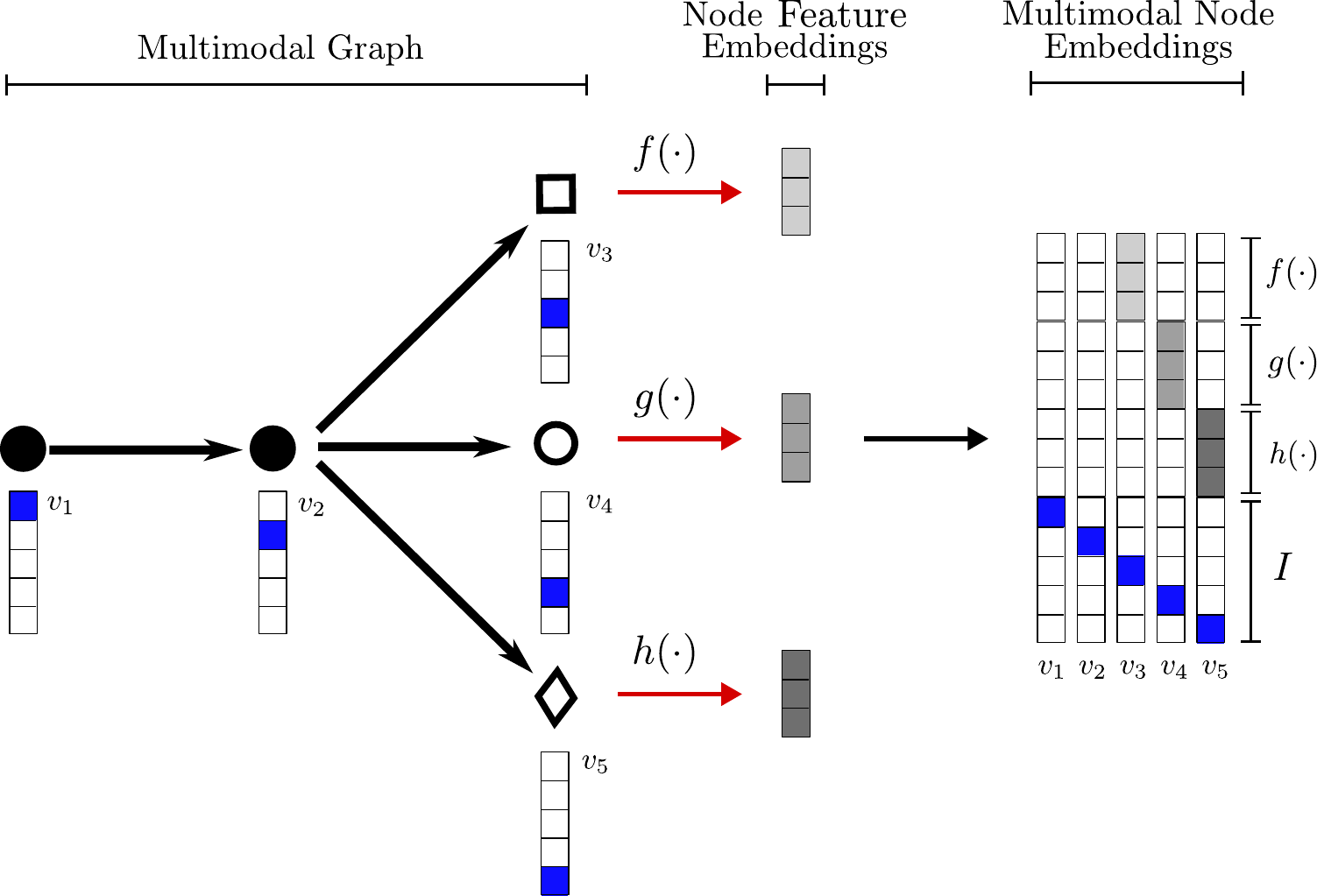}
  \caption{Overview of how our model creates multimodal node embeddings for nodes $v_1$ to $v_5$. Solid circles
	  represent entities, whereas open shapes represent literals of different modalities. The nodes' feature
	  embeddings are learned using dedicated (neural) encoders (here $f$, $g$, and $h$), and concatenated to their
  identity vectors $I$ to form multimodal node embeddings, which are fed to a message passing network.}
  \label{fig:model}
\end{figure}

We introduce our model as an extension to message passing networks which can learn end-to-end from the structure of an
arbitrary graph, and for which holds that $\bH^0=\bI$. To do so, we let $f(\cdot)$, $g(\cdot)$, and $h(\cdot)$ be feature encoders that output feature
embeddings of lengths $\ell_f$, $\ell_g$, and $\ell_h$ for nodes $v_i \in \mathcal{V}$. We define $\bF$ as the $n
\times f$ matrix of multimodal feature embeddings with $f = \ell_f + \ell_g + \ell_h$, and concatenate $\bF$ to the
identity matrix $\bI$ to form multimodal \emph{node} embeddings:

\begin{equation}
	\label{eq:input}
	\bH^0 = [\bI~\bF]
\end{equation}

\noindent of size $n \times q$ (Fig.~\ref{fig:model}).

Embedding matrix $\bH^0$ is fed together with $\bA^r$ to a message passing network, such as the R-GCN. Both encoders and
network are trained end-to-end in unison by backpropagating the error signal from the network through the encoders all
the way to the input. 

\subsection{Modality Encoders}
\label{sec:modalities}

We add encoders for five different modalities which are commonly found in knowledge graphs. We forgo discussing
relational information---the sixth modality---as that is already extensively discussed in related work on message
passing networks. For numerical and temporal information, we use straightforward deterministic encodings due to the
simplicity of the problem. For textual, visual, and spatial information we use neural encoders, for which we chose
convolutional neural networks (CNN) because of their efficiency and speed. In the case of neural encoders, we also
introduce an intermediate step in which we convert the raw values to their vector representations.

In the following, we let $\be^m_i$ be the feature embedding vector of node $v_i$ for modality $m$. The concatenation of
a node's identity vector and all its feature embedding vectors $\be^m_i$ for every $m \in \mathcal{M}$ equals the $i$-th
row of $\bH^0$.

\subsubsection{Numerical Information} 

Numerical information encompasses the set of real numbers $\mathbb{R}$, and corresponds to literal values with a
datatype declaration of \texttt{XSD:double}, \texttt{XSD:float}, and \texttt{XSD:decimal} and any subtype thereof. For
these, we simply take the values themselves as their embeddings, and represent these in a shared embedding space. We also include values
of the type \texttt{XSD:boolean} into this category due to lack of a more faithful representation, but separate their
embeddings from those of real numbers to convey a difference in semantics.

More concretely, for all nodes $v_i \in \mathcal{V}$ holds that $\be^{numerical}_i$ is the concatenation of their numerical
and boolean components, encoded by functions $f_{num}$ and $f_{bool}$, respectively. Here, $f_{num}(v_i) = v_i$ if $v_i$ is
a literal node with a value in $\mathbb{R}$. If $v_i$ is a boolean instead, we let $f_{bool}(v_i)$ be $1.0$ if $v_i$ is
\texttt{true} and $-1.0$ if $v_i$ is \texttt{false}. In both cases, we represent missing or erroneous values with $0.0$
(we assume a normalization between -1 and 1).

\subsubsection{Temporal Information}

Literal values with datatypes which follow the \textit{Seven-property model}\footnotemark such as \texttt{XSD:time},
\texttt{XSD:date} and \texttt{XSD:gMonth}, are treated as temporal information.
Different from numerical values, temporal values contain elements that are defined in a circular value space and which
should be treated as such. For example, it is incorrect to say that January and December are \emph{always} 11 months apart, as would
be implied by directly feeding the months' number to our models. Instead, it is more accurate to encode this as
\begin{equation}
	\label{eq:circ}
	f_{trig}(\phi, \psi) = [sin(\frac{2\pi \phi}{\psi}), cos(\frac{2\pi \phi}{\psi})]
\end{equation}
with $\psi$ the number of elements in the value space (here 12), $\phi$ the integer representation of the element
we want to encode, and $f_{trig}$ a trigonometric function in our encoder.

\footnotetext{https://www.w3.org/TR/xmlschema11-2}

We use this encoding for all other circular elements, such as hours ($\psi=24$) and decades ($\psi=10$). When dealing with
years however, we decided to encode smaller changes more granular than larger changes. That is, years are split into
centuries, decades, and (single) years fragments, with decades and years treated as circular elements but with centuries
as numerical values (we limit our domain to years between $-9999$ and $9999$). 

Concretely, consider date literals of the form \texttt{(+|-)YYYY-MM-DD} (we omit timezone specification for simplicity).
For every node $v_i$ of this form we let $\be^{temporal}_i$ be $1.0$ and $-1.0$ at index $j$ for \textit{CE} and \textit{BCE}, respectively,
$\be^{temporal}_i$ at index $k \neq j$ the centuries in $\mathbb{N}$, and with decades, (single) years, months, and days represented as in
Equation~\ref{eq:circ}, resulting in an embedding of length 10.

\subsubsection{Textual Information}

Vector representations for textual attributes with the datatype \texttt{XSD:string} or any subtype thereof, are created using a
character-level encoding, proposed in \cite{zhang2015character}. Hereto, we let $\bE^s$ be a $|\Omega| \times |s|$
matrix representing string $s$ using vocabulary $\Omega$, such that $\bE^s_{ij}=1.0$ if $s_j = \Omega_i$, and $0.0$
otherwise.

A character-level representation enables
our models to be language agnostic and independent of controlled vocabularies (allowing it to cope with colloquiums and
identifiers for example), as well as provide some robustness to spelling errors. It also enables us to forgo the
otherwise necessary stemming and lemmatization steps, which would remove information from the original text.
The resulting embeddings are optimized by running them through a temporal CNN $f_{char}$ with output dimension $c$, such that 
$\be^{textual}_i = f_{char}(\bE^{v_i})$ for every node $v_i$ with a textual value.

\subsubsection{Visual Information}

Images and other kinds of visual information (e.g.\ videos, which can be split in frames) can be included in a knowledge
graph by either linking to them or by expressing them as binary string literals with the datatype
\texttt{XSD:b64string} which are incorporated in the graph itself (as opposed to storing them elsewhere). In either
case, we first have to obtain the raw image files by downloading and/or converting them.

Let $im_i$ be the raw image file as linked to or encoded by node $v_i$. We can represent this image as a tensor
$\bE^{im_i}$ of size $channels \times width \times height$, which we can feed to a two-dimensional CNN $f_{im}$ with
output dimension $c$, such that $\be^{visual}_i = f_{im}(\bE^{im_i})$ for the image associated with 
node $v_i$.

\subsubsection{Spatial Information}

Spatial information includes points, polygons, and any other spatial features that consist of one or more coordinates.
These features can represent anything from real-life locations or areas to molecules or more abstract mathematical
shapes. Literals with this type of information are typically expressed using the \textit{well-known text representation}
(WKT) and therefore carry the \texttt{OGC:wktLiteral} datatype declaration. The most elementary spatial feature is a
coordinate (point geometry) in a $d$-dimensional space, expressed as \texttt{POINT}($x_1 \ldots x_d$), which can be
combined to form more complex types such as lines and polygons. 

We use the vector representations proposed in \cite{van2018deep} to represent all supported spatial features as the
enumeration of their coordinates. Let $\bE^{sf}$ be the $|\bx| \times |sf|$ matrix representation for spatial feature $sf$
consisting of $|sf|$ coordinates, and with $\bx$ the vector representation of one such coordinate. Vector $\bx$ holds
all of the coordinate's $d$ points, followed by its other information (e.g.\ whether it is part of a polygon) 
encoded as binary values. For spatial features with more than one coordinate, we also need to separate their location from
their shape to ensure that we capture both these components. To do so, we encode the location in $\mathbb{R}^d$ by taking 
the mean of all coordinates that makeup the feature. To capture the shape, we compute the global mean of all spatial
features in the graph, and subtract this from their coordinates to place their centre around the origin. 

We optimize the vector representations using a temporal CNN $f_{sf}$ with output dimension $c$, such that 
$\be^{spatial}_i  = f_{sf}(\bE^{v_i})$ for all nodes $v_i$ which express spatial features.

\section{Experiment}

We evaluate our model on an entity classification task while varying the modalities which are included in the learning
process. To do so, we compute the classification accuracies for each combination of structure and modality, as well as
all modalities combined, and evaluate this against using only the relational information and the performance
on a majority class classifier.

Another dimension that we test is how a graph's structure is represented and fed to our model, and how this influences
the performance with and without node features. The two graph representations that we test on differ only in how they
deal with literal nodes that have the same value. The most common approach is to collapse these literals into a single node, which
we will refer to as \textit{merged} literals, whereas the alternative is to keep these duplicate values separated and represent
them by as many nodes as there are values. We will call this latter configuration the \textit{split} literals.

\subsection{Implementation}

For our implementation\footnotemark, we use the R-GCN as main building block onto which we stack our various encoders.
The R-GCN can learn end-to-end on the structure of relational graphs, taking relational types into account, and which is
therefore a suitable choice to learn on knowledge graphs. If we are only interested in learning from a graph's
structure, we let $\bH^0$ be the nodes' $n \times n$ identity matrix $\bI$. (that is, $\bH^0=\bI$). To also include
literal values in the learning process, or node features, we let $\bF$ be the $n \times f$ feature embedding 
matrix and concatenate this to $\bH^0$ as in Equation~\ref{eq:input} to form $\bH^0=[\bI~\bF]$.

\footnotetext{Code available at https://gitlab.com/wxwilcke/mrgcn}

To cope with the increased complexity brought on by including node features we optimize our implementation
for sparse matrix operations by splitting up the computation of Equation~\ref{eq:rgcn} into the sum of the structural
and feature component. For this, we once again split $\bH^0$ into identity matrix $\bH_I=\bI$ and feature matrix
$\bH_F^0=\bF$, and rewrite the computation as 

\begin{equation}
	\label{eq:mrgcn}
    \bH^1 = \sigma\left(\sum_{r\in \mathcal{R}} \bA^r\bH_I\bW_I^r + \bA^r\bH_F^0\bW_F^r \right)
\end{equation}

\begin{figure}[t]
  \includegraphics[width=\linewidth]{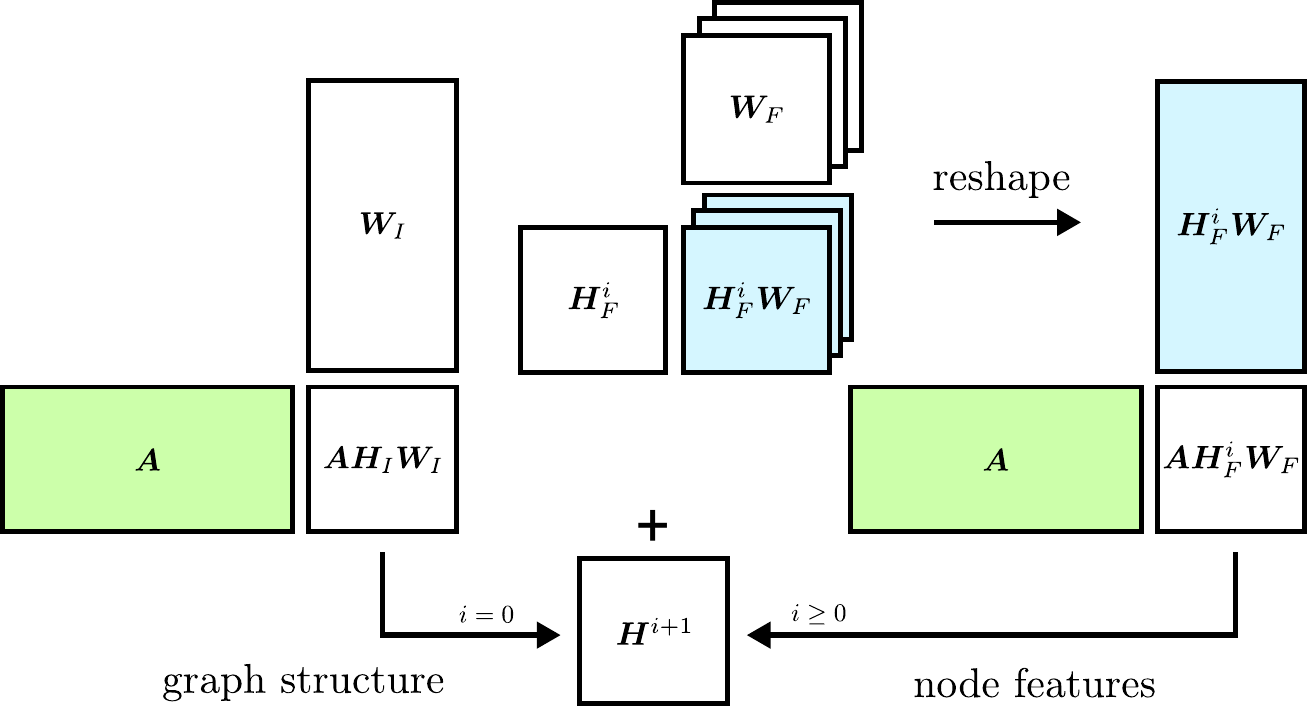}
  \caption{Graphical depiction of our implementation of Equation~\ref{eq:mrgcn}, shown as matrix operations. The
	  output of layer $i$, $\bH^{i+1}$, is computed by summing the structure and node feature
  components. If $i>0$, then $\bH^i_F = \bH^i$ and $\bA\bH_I\bW_I = 0$.}
  \label{fig:implementation}
\end{figure}

Here, $\bW_I^r$ and $\bW_F^r$ are the learnable weights for the structural and feature components, respectively. For
layers $i>0$ holds that $\bH^i_F = \bH^i$, and that $\bA^r\bH_I\bW_I^r = 0$. Note
that because $\bA^r\bH_I = \bA^r$, we can omit this calculation when computing Equation~\ref{eq:mrgcn}, and thus also no
longer need $\bH_I$ as input.
Figure~\ref{fig:implementation} illustrates this computation as matrix operations.

\subsubsection{Neural Encoders}

All three neural encoders are implemented using CNNs. For textual information, we use a temporal CNN with 4
convolutional layers, each followed by $\mathrm{ReLU}$, and 3 dense layers (Table~\ref{tab:encodertext}), which
has a minimal input sequence length of 12 characters. A similar setup is used for the spatial encoder, but with 3
convolutional layers and with a different number of filters (Table~\ref{tab:encodergeom}), and with a minimal input
length of 4 coordinates. In both cases, we trim outliers and use zero padding where needed.  For the visual encoder, we use
the efficient MobileNets architecture from~\cite{howard2017mobilenets}, with an output dimension of 128. All three CNNs
are initiated using $\mathcal{N}(0, 1)$, and are trained using mini batching (4 passes per epoch).

\begin{table}[t]
	\caption{Configuration of the textual encoder with 4 convolutional layers (top) and 3 dense layers (bottom). For
		pooling layers, \textit{Max(k/s)} lists kernel size (\textit{k}) and stride (\textit{s}), or
		\textit{Max}$(\cdot)$ when it depends on the input sequence length.}
\label{tab:encodertext}
\centering
\begin{tabular}{ccccc}
\toprule
Layer  & Filters & Kernel & Padding & Pool\\\midrule
1 & 64 & 7 & 3 & Max(2/2)\\
2 & 64 & 7 & 3 & Max(2/2)\\
3 & 64 & 7 & 3 & -\\
4 & 64 & 7 & 2 & Max($\cdot$) \\\bottomrule
\end{tabular}

\bigskip

\begin{tabular}{cc}
	\toprule
	Layer  & Dimensions\\\midrule
	5 & 256\\
	6 & 64\\
	7 & 16\\\bottomrule
\end{tabular}
\end{table}

\begin{table}[t] \caption{configuration of the spatial encoder with 3 convolutional layers (top) and 3
	dense layers (bottom). for pooling layers, \textit{max(k/s)} lists kernel size (\textit{k}) and stride (\textit{s}),
whereas \textit{avg}$(\cdot)$ depends on the input sequence length.}
\label{tab:encodergeom}
\centering
\begin{tabular}{ccccc}
\toprule
layer  & filters & kernel & padding & pool\\\midrule
1 & 16 & 5 & 2 & max(3/3)\\
2 & 32 & 5 & 2 & -\\
3 & 64 & 5 & 2 & avg($\cdot$) \\\bottomrule
\end{tabular}

\bigskip

\begin{tabular}{cc}
	\toprule
	layer  & dimensions\\\midrule
	4 & 128\\
	5 & 32\\
	6 & 16\\\bottomrule
\end{tabular}
\end{table}

The output of layer $i$ from all encoders for all nodes in $\mathcal{V}$ are concatenated to form $\bH_F^i$, which is passed to
Equation~\ref{eq:mrgcn} together with $\bA^r$. A final row-wise softmax non-linearity is added to output class probabilities. 

\subsection{Datasets}
\label{sec:datasets}

We evaluate our model on six knowledge graphs\footnotemark with different degrees of multimodality. General and
modality-specific statistics about each of these are listed in Table~\ref{tab:datasets} and~\ref{tab:datatypes},
respectively. 

AIFB, MUTAG, BGS, and AM are existing benchmark datasets for machine learning on knowledge graphs~\cite{ristoski2016collection}.
However, AIFB, BGS, and AM lack the datatype declarations needed to accurately determine the literals' modalities, which
were therefore added by us to create the AIFB+, BGS+, and AM2D+ datasets. AM2D+ further differs from AM in that we added
images, and that we pruned the graph to include only the nodes up to depth two from the labeled entities (due to the
practical difficulties caused by the increased size).

\footnotetext{Datasets available at https://gitlab.com/wxwilcke/mmkg} 

A fifth dataset, the Dutch Monument Graph (DMG), was compiled by us as benchmark for multimodal learning on knowledge
graphs, and includes information from all five modalities listed in Section~\ref{sec:modalities}, in addition to the
relational information encoded by the graph's structure. The graph integrates three existing public knowledge graphs published
by the Dutch Cultural Heritage Agency, Statistics Netherlands, and Kadaster. Together, these graphs combine various
information about monumental buildings in The Netherlands, including images and geometries. A simplified example is
given in Figure~\ref{fig:examplegraph}.

The final dataset, SYNTH, is a synthetic dataset which we created to eliminate the influence of particular
characteristics of a dataset from our results, as well as the influence of any information encoded in the graph's structure.
This allows us to ensure that any difference in results can be confidently attributed to the inclusion of a certain
modality. rather than some confounding factor. Hereto, we generated\footnotemark a random graph structure using the
Watts–Strogatz algorithm, from which we randomly sampled 256 nodes to serve as signal entities whereas the remaining nodes 
would function as noise. For each entity, we added (with $p = 0.9$) literal attributes for each of the five
modalities listed in Section~\ref{sec:modalities}, with values for our signal entities randomly drawn from two narrow
distributions (creating a binary classification problem), and with values for all other entities sampled by randomly
selecting a point in the value space of the modality at hand.

\footnotetext{Code available at https://gitlab.com/wxwilcke/graphsynth}

\begin{table*}[t]
\caption{Datasets used in our experiments. The AIFB+, AM2D+, and BGS+ datasets were extended with datatype declarations, and
images were added to the AM+ dataset. Literals with the same value are counted as the same node in the merged count, whereas
they are counted separately in the split count.}
\label{tab:datasets}
\centering
\begin{tabular}{lrrrrrr}
\toprule
Dataset            & AIFB+ & SYNTH   & DMG & MUTAG  & AM2D+ & BGS+      \\ \midrule
Facts              & 29,219 & 30,600  & 51,179 & 74,567 & 639,190 & 916,345  \\
Relations          & 45     & 38      & 37 & 23     & 123 & 103    \\
Labeled            & 176    & 256     & 600 & 340    & 1000 & 146   \\
Classes            & 4      & 2       & 6 & 2      & 11 & 2      \\ \midrule
Entities (merged)  & 6,072  & 5,976   & 25,557 & 32,621 & 174,401 & 258,519  \\
Literals (merged)  & 5,468  & 20,274  & 15,080 & 1,104    & 25,822 &  230,790  \\ \midrule
Entities (split)   & 2,835  & 4,098   & 5,704 & 22,540 & 146,609 & 103,055  \\
Literals (split)   & 8,705  & 22,152  & 34,933 & 11,185   & 53,614 & 386,254 \\ \bottomrule
\end{tabular}
\end{table*}

\begin{table*}[t]
\caption{Distribution of datatypes in the datasets. Numerical information includes all
subsets of real numbers, as well as booleans, whereas date, years, and other similar types are listed under
temporal information. Textual information includes strings and its subsets, as well as raw URIs (e.g.\ links). Images
and geometries are listed under visual and spatial information, respectively.}
\label{tab:datatypes}
\centering
\begin{tabular}{lrrrrrr}
\toprule
Dataset            & AIFB+  & SYNTH  & DMG    & MUTAG  & AM2D+     & BGS+  \\ \midrule
Numerical          & 115   & 7,382  & 1,342  & 11,185 & 11,113 & 12,332 \\
Temporal           & 1,227 & 3,727  & 219    & -      & 14,798 & 13 \\
Textual            & 7,363 & 3,701  & 29,044 & -      & 26,891 & 279,940 \\
Visual             & -     & 3,652  & 3,279  & -      & 812    & - \\
Spatial            & -     & 3,690  & 1,049  & -      & -      & 73,870 \\
Other              & -     & -      & -      & -      & -      & 20,098 \\ \bottomrule
\end{tabular}
\end{table*}

\section{Results \& Discussion}

The results are listed in Table~\ref{tab:resultsmerged} and \ref{tab:resultssplit} for merged and split literal
configurations, respectively, and report the average classification accuracy over 10 runs on the test sets. For DMG and
SYNTH, these sets were created using the 80/20/20 rule, whereas for the others we used the splits
from~\cite{ristoski2016collection}. For each dataset, we show the results for learning with and without node features,
as well as a breakdown per modality if available. All results were obtained using a two-layer R-CGN with 16 hidden
nodes, and were trained in full batch mode with Adam for 100 epochs with early stopping (after no improvement for 7
epochs) and with an initial learning rate of $0.01$. Note that all results are with respect to those of learning using
just the graph's structure, which serves as our baseline.

\begin{table*}[t]
	\caption{Entity classification results in accuracy, averaged over 10 runs, with only unique literals (merged
		configuration). \textit{Structure} uses only the relation information whereas 
		\textit{Structure + Features} also includes information from all supported modalities. The rest provides a
		breakdown per modality.}
\label{tab:resultsmerged}
\centering
\begin{tabular}{lrrrrrr}
\toprule
Dataset               & AIFB+  & SYNTH   & DMG    & MUTAG  & AM2D+    & BGS+  \\ \midrule
Majority Class        & 0.4167 & 0.5000  & 0.1667 & 0.6618 & 0.3333   & 0.6552\\
Structure             & \bf{0.9583} & 0.6942  & \bf{0.5917} & 0.6956 & \bf{0.8803} & 0.8242 \\
Structure + Features  & 0.8861 & \bf{0.8173}  & 0.5317 & \bf{0.7324} & 0.8399 & \bf{0.8414} \\ \midrule 
Structure + Numerical & 0.9583 & 0.7500  & 0.5958 & \bf{0.7324} & 0.8773 & \bf{0.8414} \\
Structure + Temporal  & \bf{0.9666} & 0.7981  & 0.5767 & -      & \bf{0.8694} & 0.8242 \\
Structure + Textual   & 0.9139 & 0.6952  & \bf{0.7317} & -      & 0.8152 & 0.8276 \\
Structure + Visual    & -      & \bf{0.9250}  & 0.4042 & -      & 0.8187 & - \\
Structure + Spatial   & -      & 0.4962  & 0.6233 & -      & -      & 0.7552 \\\bottomrule
\end{tabular}
\end{table*}

\begin{table*}[t]
	\caption{Entity classification results in accuracy, averaged over 10 runs, with literals with the same value are
		kept separately (split
	configuration). \textit{Structure} uses only the relation information whereas 
		\textit{Structure + Features} also includes information from all supported modalities. The rest provides a
		breakdown per modality. }
\label{tab:resultssplit}
\centering
\begin{tabular}{lrrrrrr}
\toprule
Dataset               & AIFB+  & SYNTH   & DMG    & MUTAG  & AM2D+     & BGS+  \\ \midrule
Majority Class        & 0.4167 & 0.5000  & 0.1667 & 0.6618 & 0.3333   & 0.6552\\
Structure             & \bf{0.9167} & 0.5019  & 0.2308 & 0.6559 & \bf{0.8561} & 0.8414 \\
Structure + Features  & 0.8611 & \bf{0.7462}  & \bf{0.4850} & \bf{0.6721} & 0.8394 & \bf{0.8449} \\ \midrule
Structure + Numerical & 0.9167 & 0.6558  & 0.3096 & \bf{0.6721} & 0.8583 & \bf{0.8414} \\
Structure + Temporal  & 0.9167 & 0.7442  & 0.2350 & -      & \bf{0.8593} & 0.8380 \\
Structure + Textual   & 0.7611 & 0.7039  & \bf{0.5456} & -      & 0.8182 & 0.8311 \\
Structure + Visual    & -      & \bf{0.8981}  & 0.2508 & -      & 0.8515 & - \\
Structure + Spatial   & -      & 0.4558  & 0.2433 & -      & -      & 0.8276 \\ \bottomrule
\end{tabular}
\end{table*}

The overall results show a slight to considerate improvement when including certain node features for almost all
datasets, except for AM2D+ of which the difference might also be due to randomness in initialization. However, any difference
in performance seems to depend strongly on which modality we include: some modalities improve the baseline by little to
nothing, whereas others improve or worsen it considerably. These differences appear to stack when we include information
from all supported modalities, with the best or worst modality pulling or dragging the combined results up or down,
respectively.  The results on SYNTH seem to indicate that at least part of this is caused by the particular
characteristics of the datasets, such as noisy signals from semi-random or task-irrelevant attributes (e.g.\
identifiers). However, our models should learn to ignore these signals, which might indicate another cause, for instance
because the weights of the neural encoders are shared for all attributes of the same modality despite the different
domains of properties that makeup that modality.

We can also see a difference depending on whether or not we merge literal values, with an overall similar or lower
performance when we split literals. The results on SYNTH seem to indicate that this difference---$0.69$ vs.  $0.50$ on
baseline---can be attributed to information from literal values being encoded in the graph's structure, which suggests
that explicitly including additional modalities may not always be worth the increased complexity as the information is already
implicitly present in the graph. This is supported by the
difference between the majority class and the baseline results, which shows how much of the signal exactly is captured in the
relational information. Nevertheless, on datasets where only little or no signal is present in the structure,
such as SYNTH and DMG, including information from other modalities appears to increase the performance significantly
(except for MUTAG, which might be caused by the low ratio of literals to entities). 

Finally, we must note the poor results with spatial features on SYNTH, despite providing a slight gain and drop in
performance for DMG and BGS, respectively. As the results on SYNTH are even below that of the majority class, we believe
that this is a problem of how we generated the spatial features, rather than an indication that including spatial
information should be avoided.


\section{Discussion}

In this work, we have introduced a model for end-to-end multimodal learning on heterogeneous knowledge graphs which
treats literals as first-class citizen by encoding them in accordance with the characteristics of their modalities. 

Our results indicate that, overall, including information from other modalities can improve the performance of our
models either slightly or considerably, depending strongly on the characteristics of the data and whether or not we
merge literals with the same value (thereby encoding literal information in the structure of the graph). We believe that
these results support our supposition that by including as much information as possible, staying closer to the original
and complete information in the graph, enables our models to learn better internal representations of its nodes, with an
overall increase in performance as result. Nevertheless, more research is needed to understand how we can best include
multimodal node features in the learning process.

\subsection{Limitations and future work}

Our aim has currently been to demonstrate that we can train a multimodal message passing model end-to-end which can
exploit the  information contained in a graph's literals and naturally combine this with its relational counterpart,
rather than to established that our approach reaches state-of-the-art performance, or even to measure its performance
relative to other published models.

To properly establish which type of model architecture performs best in multimodal settings, and whether message passing
models provide an advantage over more shallow embedding models without message passing, we require more extensive,
high-quality, standard benchmark datasets with well-defined semantics (i.e.\ datatype and/or relation range
declarations). To this end, we have contributed the DMG knowledge graph, which contains literal nodes from various
modalities. We have also contributed variants of three existing benchmark datasets, which we made suitable for
multimodal learning by adding the necessary datatype declarations (and in one case also images). Nevertheless, to determine
precisely what kind of data is most fitting for this form of learning we are likely to require an iterative
process where each generation of models provides inspiration for the next generation of benchmark datasets and vice versa.

We also note that all knowledge graphs currently used in entity classification (including DMG) have a limited amount of
labeled entities. This means that there is likely a large amount of variance in the estimated accuracies. To robustly
establish best practices for model architectures, benchmark datasets with test sets of at least 10\,000 entities will
eventually be required. Our effort to generate a synthetic benchmark dataset which we can tweak as we wish might be a
step in the right direction, although real-life data is much preferred. 
 
So far we have only tested our method on entity classification, as that is where the message passing aspect of the R-GCN
seems to make the most difference. It is yet to be established whether this approach can also yield a performance
benefit in the setting of link prediction---a question we are currently exploring. Note that any score function from the
embedding methods in \cite{gesese2019survey} can, in principle, be combined with the R-GCN layers and the encoders. This
leads to a large configuration space of possible models. We reiterate that the first thing that is required to explore
this space effectively is high-quality benchmark data.

We performed little hyperparameter tuning in our model development, since the aim was not to tune a model to optimal
performance, but only to show that information from literal nodes could, in principle, benefit performance. Due to the
exploratory nature of the project, the test set was used for evaluation multiple times during development. In future
work, a more rigorous protocol will be followed, to avoid the effects of multiple testing.

Future work will also investigate the trade off between potentially improving the performance by using a separate set of
learnable weights per relation (as opposed to sharing weights amongst literals of the same modality) and the
complexity this would add. Another promising angle is to explore techniques to reduce the overall complexity of a
multimodal model: the necessity of full batch learning with many message passing networks---a known limitation---makes it
challenging to learn from large graphs; a problem which becomes even more evident as we start adding multimodal node features. 

Lastly, a promising direction of research is the use of pretrained encoders. In our experiments, we show that the encoders
receive enough of a signal from the downstream network to learn a useful embedding, but this signal is complicated by
the message passing head of the network, and the limited amount of data. Using a modality-specific, pretrained encoder,
such as GPT-2 for language data~\cite{radford2019language} or Inception-v4 for image data~\cite{szegedy2017inception},
may provide us with good general-purpose feature at the start of training, which can then be fine-tuned to the specifics
of the domain.

\subsection{Conclusion}

Learning end-to-end on heterogeneous data has a lot of promise which we have only scratched the surface of. A model
that learns in a purely data-driven way to use information from different modalities, and to integrate such information
along known relations, has the potential to allow practitioners a much greater degree of hands-free machine learning on
multimodal heterogeneous data. 

We hope that our proof-of-concept serves as an inspiration, and will lead not only to further experimentation in model
configurations, but also to the development of larger and even higher-quality benchmark datasets which are reflective of
real-world use cases.

\section*{Acknowledgments}

We express our gratitude to Wouter Beek from Triply\footnote{https://www.triply.cc} for helping us obtain two of the
three knowledge graphs that make up the Dutch Monument Graph. This project is supported by the NWO Startimpuls programme
(VWData - 400.17.605) and by the Amsterdam Academic Alliance Data Science (AAA-DS) Program Award to the UvA and VU
Universities.

\bibliographystyle{plain}
\bibliography{bibliography}

\end{document}